\documentclass{article}
\usepackage{spconf,amsmath,graphicx}
\usepackage{hyperref}
\usepackage{url}
\usepackage{spconf,amsmath,graphicx}
\usepackage{booktabs}
\usepackage[shortlabels]{enumitem}
\usepackage{subcaption}
\usepackage{MnSymbol}
\usepackage{wasysym}
\usepackage{hyperref,url,latexsym,color,soul}
\usepackage{makecell}
\usepackage{verbatim}

\makeatletter
\def\blfootnote{\gdef\@thefnmark{}\@footnotetext}
\makeatother

\title{Using VAEs and Normalizing Flows for One-shot Text-To-Speech Synthesis of Expressive Speech}
\name{Vatsal Aggarwal, Marius Cotescu, Nishant Prateek, Jaime Lorenzo-Trueba, and Roberto Barra-Chicote}
\address{Amazon, Cambridge UK}
\begin{document}
%
\maketitle
\begin{abstract}
We propose a Text-to-Speech method to create an unseen expressive style using one utterance of expressive speech of around one second. Specifically, we enhance the disentanglement capabilities of a state-of-the-art sequence-to-sequence based system with a Variational AutoEncoder (VAE) and a Householder Flow. The proposed system provides a $22\%$ KL-divergence reduction while jointly improving perceptual metrics over state-of-the-art. At synthesis time we use one example of expressive style as a reference input to the encoder for generating any text in the desired style. Perceptual MUSHRA evaluations show that we can create a voice with a $9\%$ relative naturalness improvement over standard Neural Text-to-Speech, while also improving the perceived emotional intensity ($59$ compared to the $55$ of neutral speech). 
\end{abstract}
\begin{keywords}
Text-to-speech, data efficiency, and semi-supervised learning
\end{keywords}

\section{Introduction}
\label{sec:intro}
Neural Text-to-Speech models, such as Tacotron and WaveNet \cite{oord2016wavenet, shen2018natural}, are able to produce high quality speech with naturalness close to that of real humans. However, these models require several hours of data to produce high-quality voices.

This work focuses on creating high-quality expressive speech using only one utterance of a target style without requiring retraining. We show that we are able to create high-quality expressive voices at low cost and improve the naturalness of our baseline \textit{state-of-the-art} neutral text-to-speech system. Our approach relies on a architecture combining Householder Flows and Variational AutoEncoders (VAE) in a novel way that provides better KL-Divergence (KLD) between an isotropic Gaussian prior and our VAE latent variables, alongside better reconstruction than existing methods. Reductions in KLD provide lower correlation between our latent variables. Consequently, our approach provides improved manipulation of the latents  \cite{hsu2018hierarchical}, substantial improvements in the disentanglement of the latent space, and raises perceptual metrics of synthesized expressive speech whilst using only one reference utterance.

To the authors knowledge, this work is the first to tackle the problem of one-shot synthesis of an unseen expressive style. Although there is existing work in the related fields of prosody generation \cite{skerry2018towards} and prosody transfer \cite{akuzawa2018expressive}, these approaches rely on either modelling expressive speech or transferring a seen expressive style to an unseen speaker, and are trained on a large corpora of expressive speech. In contrast, we use only one utterance to create an unseen style. Our problem is also distinct from that of speaker-adaptation \cite{nachmani2018fitting, jia2018transfer}, as we wish to improve expressiveness by including emotion for existing speakers, not learn entirely new speakers. Existing work on emotional speech focuses on explicit emotional speech models \cite{lorenzo2018investigating}, model adaptation \cite{wu2018rapid} or voice conversion \cite{gao2018nonparallel,shankar2019multi}, which again require significantly more data than the single utterance required by our proposed system.

\blfootnote{Corresponding author email: agvatsal@amazon.com.}

\vspace{-2mm}
\section{Proposed Approach}
\label{sec:approach}
\begin{figure}[t]
	\centering
	\includegraphics[width=1\columnwidth]{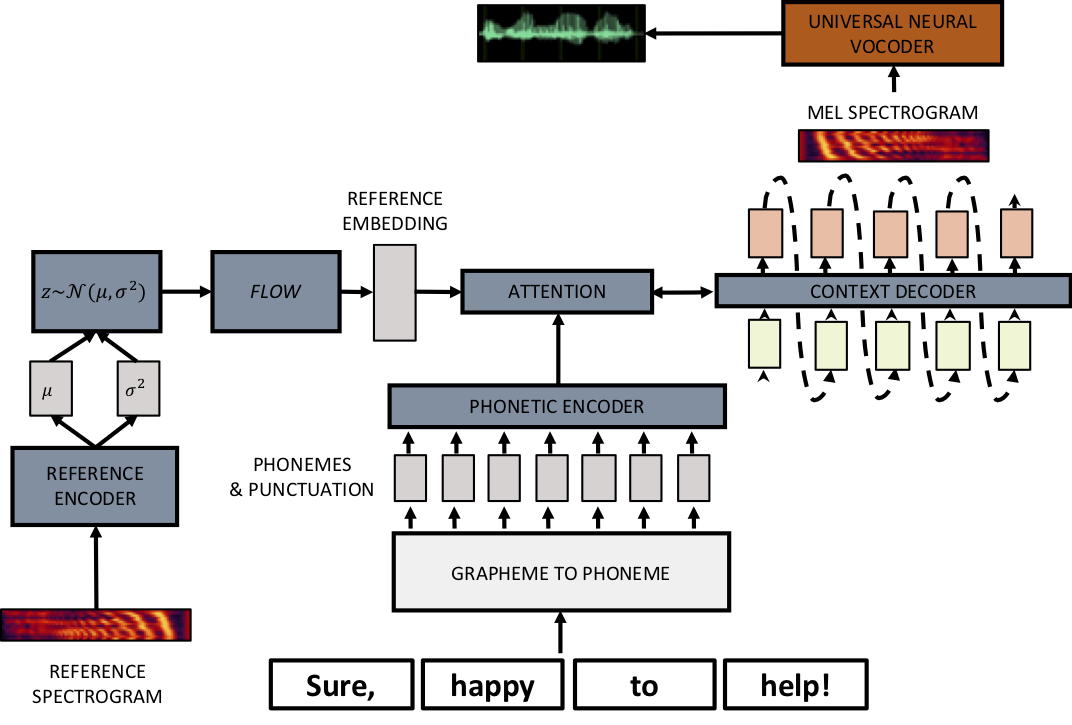}
	\caption{Overview of the system's architecture: a Variational seq-to-seq auto-encoder conditioned on the phonetic sequence. The Universal Neural Vocoder translates the generated spectrogram into the final speech signal.}
	\label{fig:Architecture}
\end{figure}
For our proposed architecture, we extend a \textit{state-of-art} sequence-to-sequence NTTS \cite{shen2018natural,prateek2019other,latorre2019effect} system. This base seq2seq model uses an attention module to map the concatenated outputs of an  RNN-based Phoneme Encoder and a mel-spectrogram conditioned Reference Encoder to mel-spectrograms via an autoregressive  decoder. The reference encoder is trained as the aggregate posterior of a VAE \cite{kingma2013auto}, and is conditioned on mel-spectrograms to predict the mean and variance of a 64-dimensional diagonal Gaussian Distribution (with similar architecture to \cite{skerry2018towards}). Thus, the reference encoder and the decoder form a text-conditioned VAE, with the Phoneme Encoder providing text-conditioning to our network. We consider this as our baseline system, referred to as Vanilla VAE.

\begin{figure}[t]
	\centering
	\includegraphics[width=1\columnwidth]{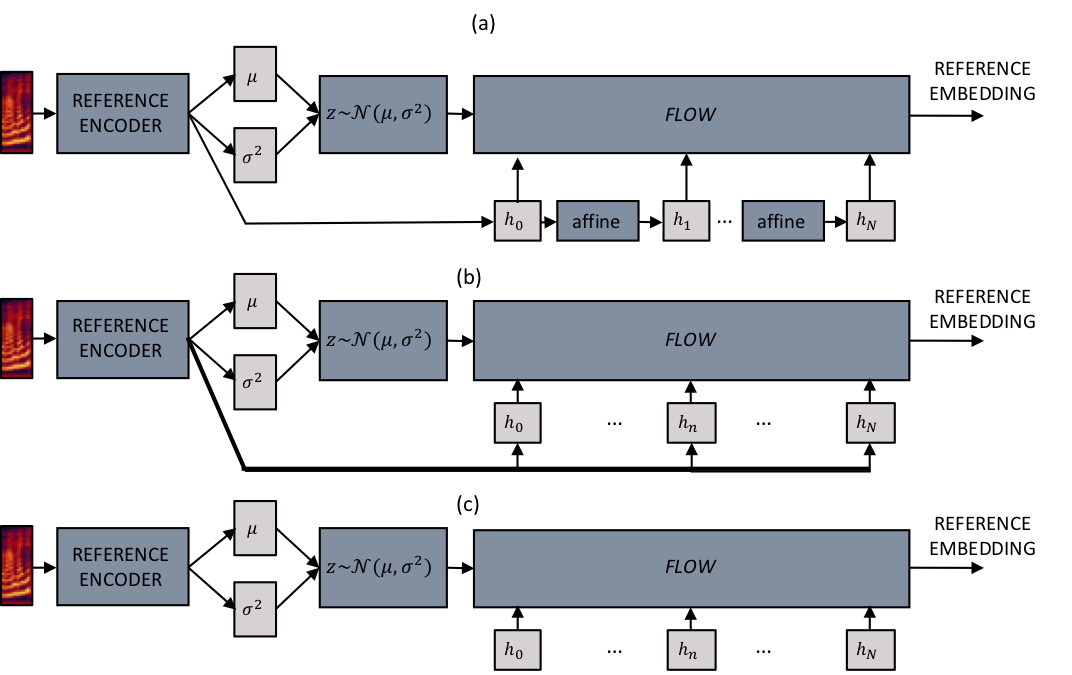}
	\caption{Diagram of proposed architectures: Architecture-1 (a), Architecture-2 (b), and Architecture-3 (c). $h_i$ refers to the $i^{th}$ householder vector.}
\label{fig:vae_plus_flows_architectures}
\end{figure}

Our proposed system extends this base model by adding a Householder Normalizing Flow \cite{tomczak2016improving} (as presented in Figure~\ref{fig:Architecture}) to the text-conditioned VAE. Householder Flows consist of multiple steps of easily invertible affine transformations that transform samples from a diagonal Gaussian to samples from a full co-variance Gaussian. We hypothesise that Householder Normalizing Flows allow samples from the reference encoder to better match the diagonal normal Gaussian prior while allowing the samples after the flow to become more co-related. We believe that this allows for both improved disentanglement and better reconstruction by the decoder. To investigate this claim, we tried a variety of architectures differing only in how the Householder vectors used to define each step of the householder flow are predicted. We now summarize our different architectures (see also Figure \ref{fig:vae_plus_flows_architectures}). 
\begin{itemize}
    \item Architecture-1: We follow the original implementation of Householder Flows \cite{tomczak2016improving} - predicting the first householder vector as an additional output of the reference encoder, with all other vectors predicted via subsequent affine transformations. 
    \vspace{-2.5mm}
    \item Architecture-2: We predict all the Householder vectors as separate outputs of the reference encoder.
    \vspace{-2.5mm}
    \item Architecture-3: We propose a novel architecture where Householder vectors are globally shared across all utterances. This architecture is distinct from all existing implementations of Householder Flows, as the vectors are not conditioned on the input to the VAE encoder in any way (e.g. Architecture-1). 
\end{itemize}
Each architecture is evaluated using 2, 4, 8 and 16 householder vectors as hyperparameters. Across all our architectures, samples from the output of the flow are concatenated with the output of the phoneme encoder instead of samples from the output distribution of the reference encoder. 

During training, we use a large training corpus (Section \ref{sec:experiments}) and feed the target spectrogram into the reference encoder. At synthesis time, we instead use a spectrogram from the 1-shot data-set (also described in Section \ref{sec:experiments}) as input to the reference encoder and feed phonemes of the target text into the phoneme encoder. To complete the TTS pipeline, we use the vocoder of  \cite{lorenzo2019towards} to convert the mel-spectrograms to audio waveforms.

\section{CORPORA}
\label{sec:experiments}
\vspace{-1mm}
We train our models on a training data-set combining two corpora: a high-quality proprietary multi-speaker corpus, and a subset of the VCTK \cite{veaux2017cstr} corpus. Both data-sets contained natural speech waveforms from English speakers speaking in non emotional styles. Our internal corpus contains 181 hours of speech from 13 speakers (each contributing between 5 and 35 hours), which we combine with 21 English VCTK speakers (each with 23 minutes of recordings). 

Alongside this training corpus, we also have access to a small corpus of expressive recordings for one of our internal speakers. This speaker exhibits the `excited' emotion across three intensity levels: low, medium and high. We randomly picked one utterance for each intensity level to condition the reference encoder in our 1-shot experiments. Furthermore, we selected $50$ text prompts to be synthesized for perceptual evaluations, ensuring that each prompt corresponds to an appropriate emotional and intensity level given its content.

\vspace{-3mm}
\section{RESULTS}
\vspace{-1mm}
\label{sec:experiments}
We evaluated each system using both objective and perceptual metrics. Our objective metrics are the KL-divergence of our reference encoder and the final teacher-forced L2 loss achieved by our the decoder. To choose systems for perceptual evaluations, we compare our system against other \textit{state-of-the-art} methods with HouseHolder Flows (across multiple hyper-parameter configurations),  and choose the best model with and best model without flows. Our perceptual evaluations target speech naturalness, emotional intensity and signal quality.
\begin{figure}[t]
	\centering
	\includegraphics[width=\columnwidth]{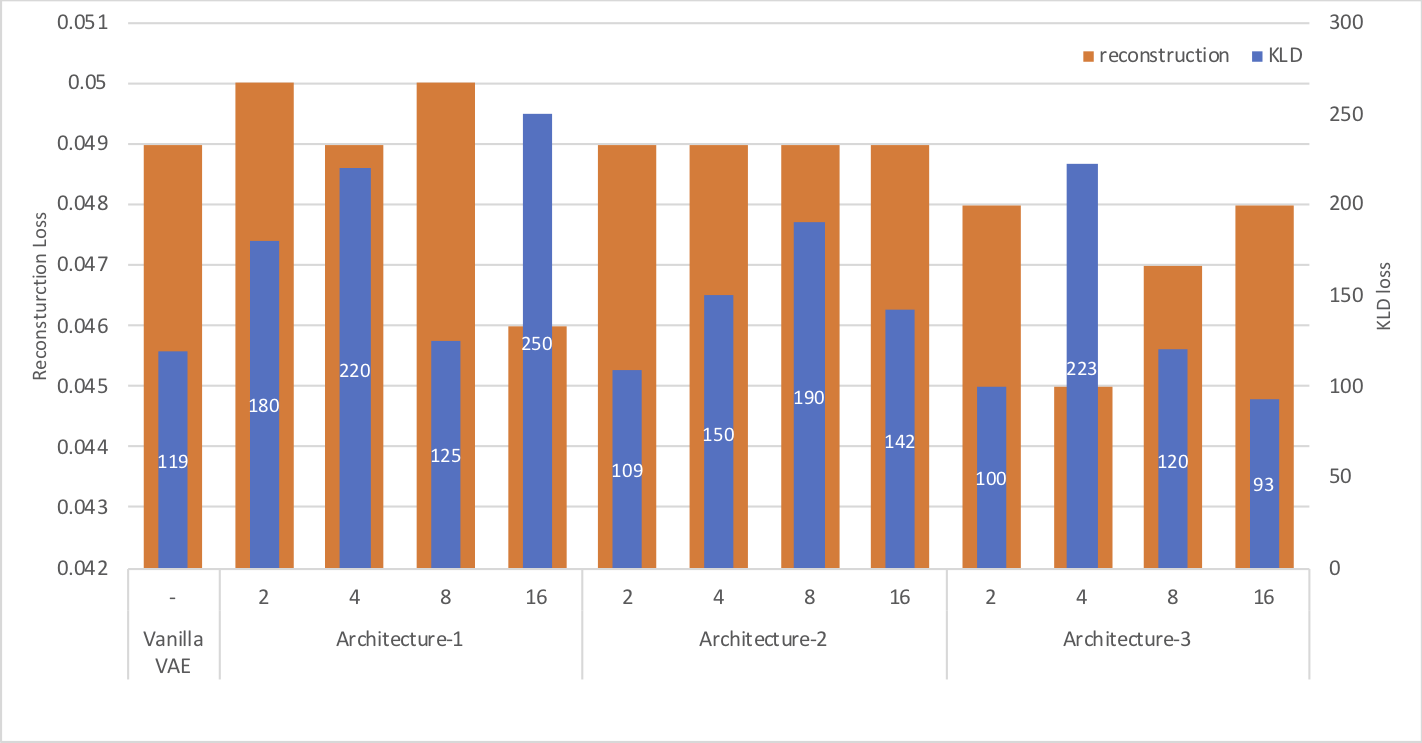}
	\caption{KL-Divergence (blue bars) and Reconstruction Objective (orange bars) Metrics. We plot the average loss for the final epoch when training is stopped for each architecture and hyper-parameter presented in the Proposed Approach section.}
	\label{fig:obj}
\end{figure}

\subsection{Objective Evaluation}
\label{ssec:obj}
We compare the Vanilla VAE system summarized in the Section 2, with the three other architectures predicting householder vectors in different ways. Our proposed architecture, Architecture-3, achieves the lowest combination of KL-divergence and Reconstruction Loss across all systems (Figure \ref{fig:obj}). Furthermore, it is the only architecture which provides lower KL-divergence, reducing it by $22$\%, or improved disentanglement, while also providing better reconstruction when comparing against Vanilla VAE. From informal listening, this model indeed provided better quality in limited scenarios as compared to the other models, with flows, that we tried. We believe this is because the flow in our proposed architecture is now able to learn transformations of factors that are relevant for speech in general rather than having to learn how to transform factors that could change with different inputs. We will refer to this architecture with the best hyper-parameter (16 householder vectors) as VAE+FLOW for the rest of this paper.

In the next section, we verify that the superior disentanglement achieved by our model (without a loss in reconstruction loss), allows for convincing 1-shot generalization of expressive speech.

\vspace{-3mm}
\subsection{Perceptual Evaluations}
\label{ssec:perc}
To evaluate our model with respect to naturalness, emotional strength, and signal quality, we ran three MUltiple Stimuli with Hidden Reference and Anchor (MUSHRA) tests \cite{recommendation2001bs} across 40 listeners from Amazon Mechanical Turk. 

For naturalness, we asked our listeners ``Please rate the naturalness of the systems. Make sure that at least 1 system per screen is always rated 100." and provided a scale from 0 (Very poor) to 100 (Completely natural). To measure emotional strength, we asked ``Please rate the strength of the emotion of the systems. Consider the neutral sounding sample (there is one provided for each utterance) as 0 and an extremely emotional sounding sample as 100." and provided a scale from 0 (Neutral) to 100 (Very Emotional). Finally, to measure signal quality, we asked  ``Please rate the systems in terms of audio quality. Try to ignore the content of the speech and the expressivity and instead focus on the quality of the audio signal (e.g glitches, clicks, noise...)." and provided a scale from 0 (Very poor) to 100 (Very good). 

Each MUSHRA consisted of 4 systems: recordings of Emotional Speech (hidden higher anchor), Neutral TTS (hidden lower anchor), Vanilla VAE and VAE+FLOW. We show each listener 50 utterances with excited emotions with prompts as described in the Corpus section. To measure statistical significance, we ran t-tests with Bonferroni-Holm corrections at a $95$\% significance level. Our null-hypothesis is that the mean MUSHRA scores of the listening population for the two systems is equal.

\subsubsection{Naturalness}
\label{sssec:natural}
\begin{figure}
	\centering
	\includegraphics[width=\columnwidth]{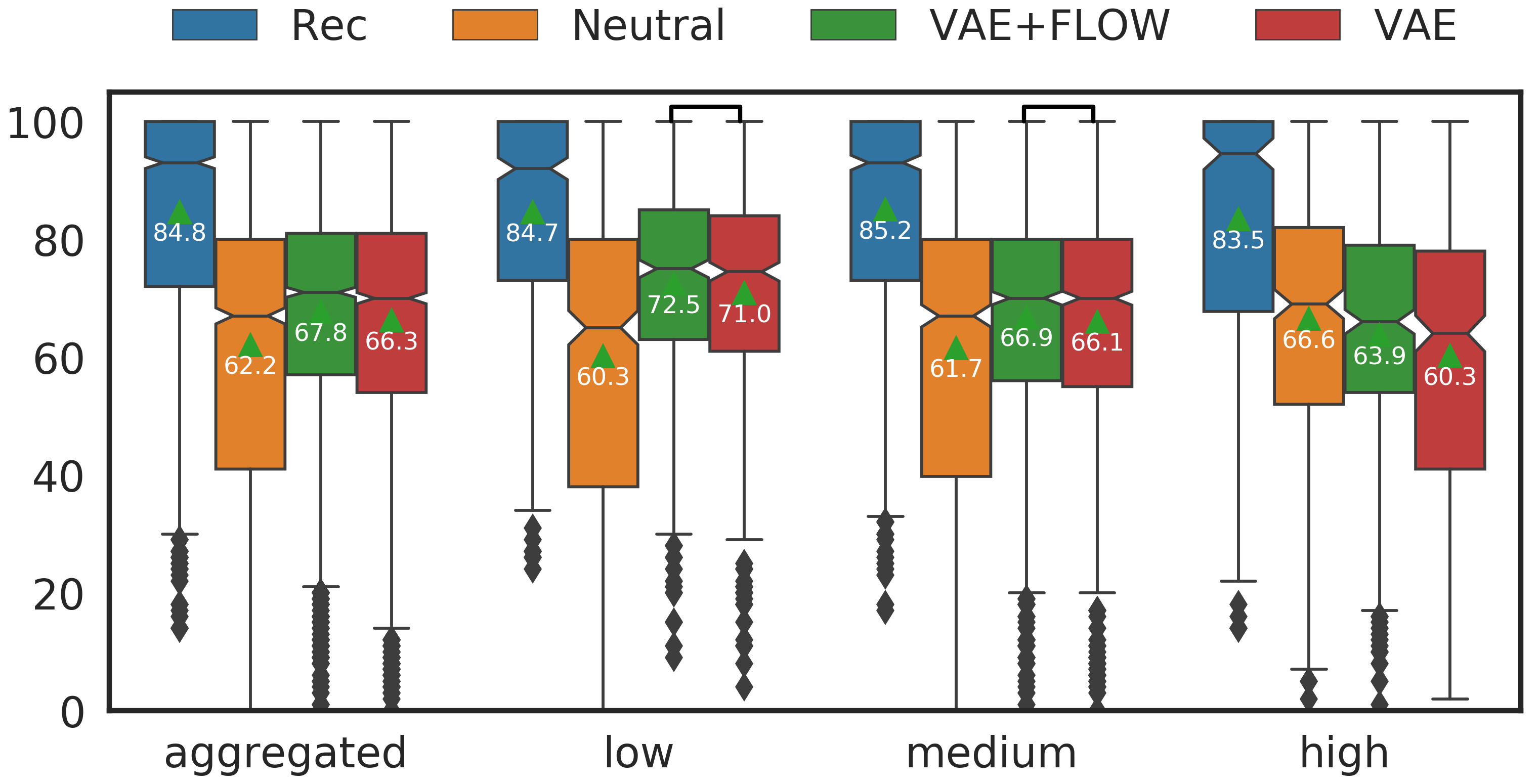}
	\caption{Perceptual Results of naturalness for low, medium, high and aggregated emotional intensities. We plot MUSHRA responses of listeners as box-plots for each intensity. The number in white and location of the green triangle represents the mean of the listeners response for that system.}
	\label{aggregrated2}
\end{figure}

MUSHRA evaluations of naturalness are presented in Figure~\ref{aggregrated2}. When aggregated across all emotional intensities, there is a statistically significant difference between all pairs of systems. Both 1-shot based expressive systems (VAE+FLOW and VAE) obtain higher naturalness ($66.3$ and $67.8$ respectively) than the neutral system ($62.2$). When we break down the results per intensity conveyed in the reference utterance, this improvement in naturalness of both the 1-shot based expressive systems with respect to the neutral system is maintained for low and medium intensities. However, the difference between the two 1-shot expressive systems is not statistically significant. In contrast, when using high intensity utterances as the reference, there is a statistically significant degradation respect to neutral ($66.6$) for both expressive systems: partially reduced by VAE+FLOW ($63.9$) as compared to the vanilla VAE ($60.3$). 

\subsubsection{Emotional Strength}
\label{sssec:emo-inten}
\begin{figure}
	\centering
	\includegraphics[width=\columnwidth]{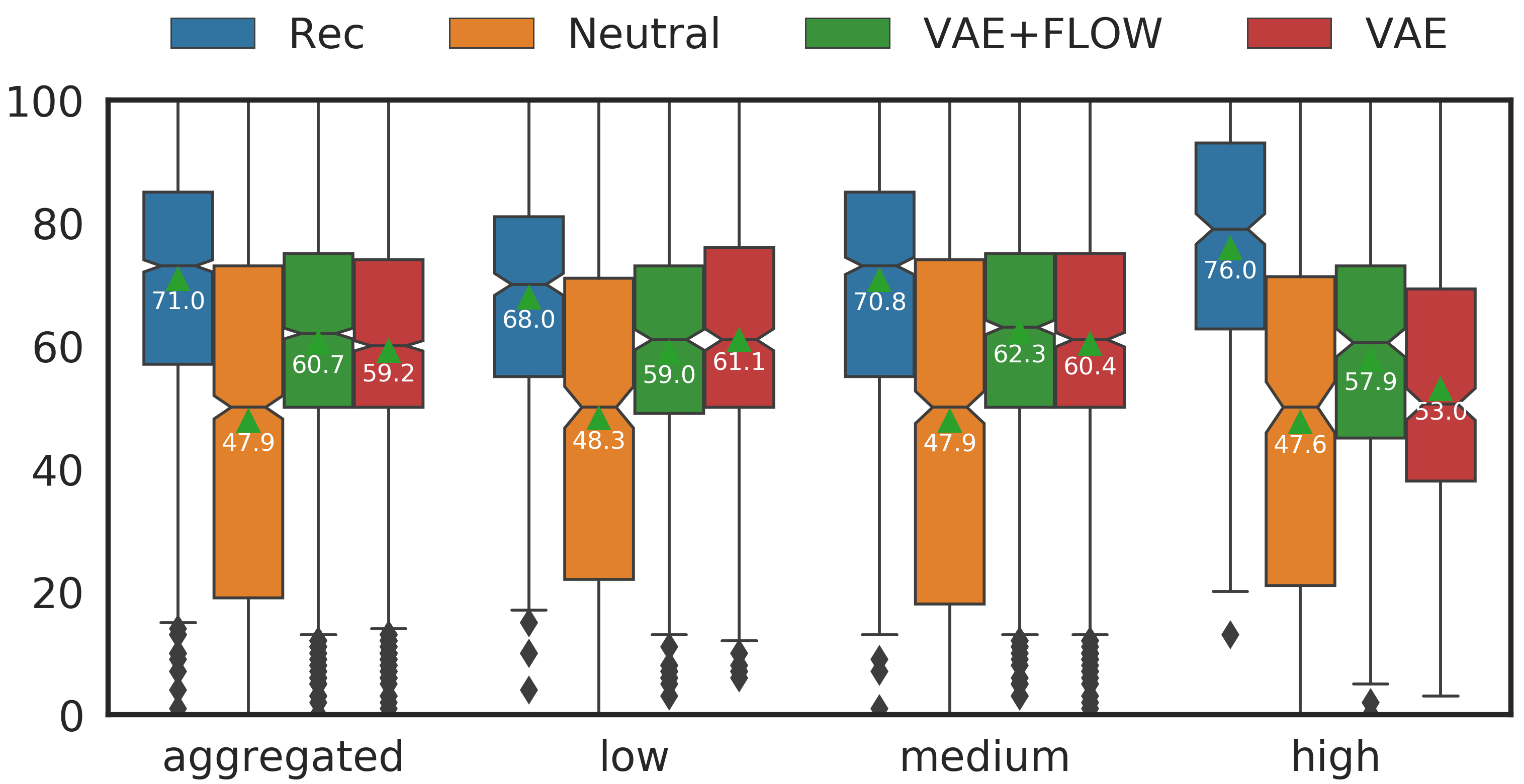}
	\caption{Perceptual Results for emotional strength across low, medium, high and aggregated emotional intensities. Legend and metrics are reported as in Figure~\ref{aggregrated2}.}
	\label{aggregrated}
\end{figure}

For emotional strength (Figure \ref{aggregrated}), we found significant differences between all pairs of systems, in the aggregated results as well as results split by intensity. As expected, listeners scored the emotional strength of the neutral system consistently (and lower than rest of the systems) across all the evaluations and both the 1-shot expressive systems were scored with a higher mean score than neutral.

\subsubsection{Signal Quality}
\begin{figure}
	\centering
	\includegraphics[width=\columnwidth]{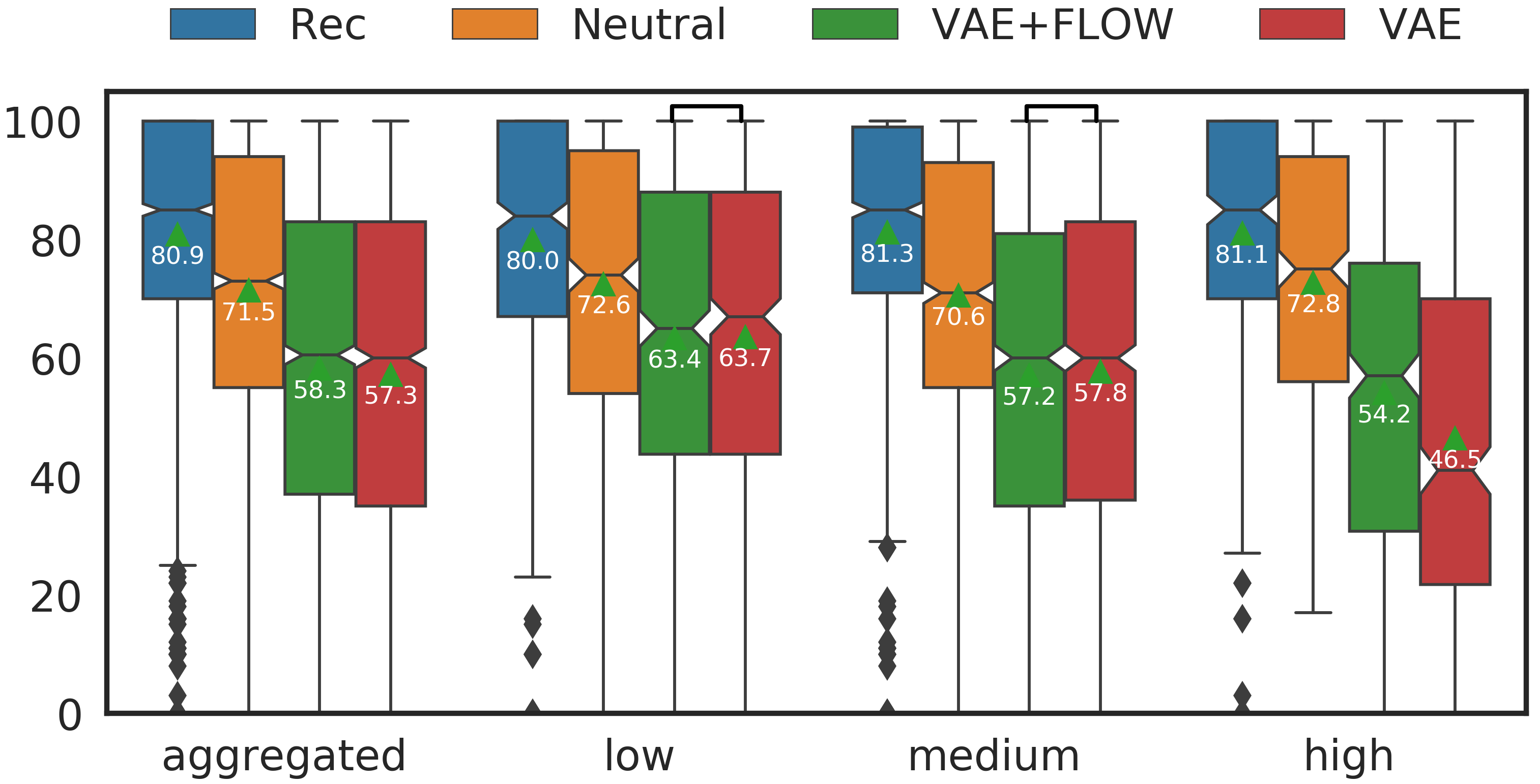}
	\caption{Perceptual Results for signal quality for low, medium, high and aggregated emotional intensities. Legend and metrics are reported as in Figure~\ref{aggregrated2}.}
	\label{aggregrated3}
\end{figure}

Figure \ref{aggregrated3} shows the perceptual evaluations of signal quality. When aggregated across all emotional intensities, we see a significant difference between all pairs of system. Furthermore, for low and medium intensities the difference between flow and no-flow is not significant, whereas differences between all other system pairs are significant. In the high intensity case, the difference between all pairs of systems is significant.

\vspace{-4mm}
\section{DISCUSSION}
\vspace{-1mm}
Overall, both Vanilla VAE and VAE+FLOW provide a method to perform one-shot adaptation of a new expressive style. Vanilla VAE provided a statistically significantly improvement in emotional strength ($23\%$ relative), and naturalness ($2\%$ relative) over Neutral TTS despite listeners detecting a degradation in signal quality ($20\%$ relative). Our proposed VAE+FLOW architecture, leveraging a simple computationally inexpensive flow, is able to statistically significantly improve further upon Vanilla VAE in emotional strength ($2.5\%$ relative) naturalness ($2.2\%$ relative), and signal quality ($1.7\%$ relative).

After breaking down the perceptual results by emotional intensity, we found that the relative improvement achieved by VAE+FLOW over Vanilla VAE increases with emotional intensity. In this scenario, a higher intensity means a bigger acoustic divergence from the neutral speech. We, therefore, hypothesize that our proposed VAE+FLOW architecture performs better at the extreme cases of high intensity due to improved disentanglement, leading to better generalization capabilities. 

This hypothesis is supported by our perceptual tests on emotional strength. All systems suffer a drop in naturalness when synthesizing high excitement (due to the degradation in signal quality caused by trying to synthesize speech very different to the training corpus). However, the improvement in generalization by the VAE+FLOW system provides a smaller naturalness degradation, compared to Neutral TTS, than the Vanilla VAE system in this case ($4\%$ and $9.5\%$ respectively). Finally, we note that while we're consistently increasing emotional strength in the low and medium case, these results are not translating proportionately to listeners perception of naturalness despite natural recordings of those emotions being rated highly. We hypothesize this is due to the limited signal quality that 1-shot systems can currently achieve.
\vspace{-2.5mm}
\section{CONCLUSIONS}
\vspace{-1.5mm}
We proposed a novel approach using householder flows within a text-conditioned VAE that overcomes the common trade-off between disentanglement and reconstruction in a state-of-the-art VAE implementation. The proposed flow-based VAE system significantly reduced KL-divergence ($22\%$ relative reduction) whilst providing a $2\%$ relative improvement in reconstruction. Furthermore, we showed that pre-existing methods can perform 1-shot adaption to a new expressive style with only one utterance. Our VAE+FLOW model improved naturalness and emotional strength over an existing state-of-the-art system particularly when generalizing across more extreme prosody differences (with respect to the neutral training data) using only one utterance.

We have shown that signal quality is the bottleneck preventing joint improvement of naturalness and emotional strength. Future investigations are necessary to create voices with limited resources across a wider variety of expressive styles, seeking to further improve naturalness without harming expressiveness.

\vspace{-1.5mm}
\section{Acknowledgment}
\vspace{-2.5mm}
We would like to thank Pablo Garcia Moreno, Jasha Droppo, Henry Moss, Andrew Breen and Thomas Drugman for insightful discussions relevant to this work.

\vfill\pagebreak
\bibliographystyle{IEEEbib}
\bibliography{strings,refs}

\begin{thebibliography}{10}

\bibitem{oord2016wavenet}
Aaron van~den Oord, Sander Dieleman, Heiga Zen, Karen Simonyan, Oriol Vinyals,
  Alex Graves, Nal Kalchbrenner, Andrew Senior, and Koray Kavukcuoglu,
\newblock ``Wavenet: A generative model for raw audio,''
\newblock {\em arXiv preprint arXiv:1609.03499}, 2016.

\bibitem{shen2018natural}
Jonathan Shen, Ruoming Pang, Ron~J Weiss, Mike Schuster, Navdeep Jaitly,
  Zongheng Yang, Zhifeng Chen, Yu~Zhang, Yuxuan Wang, Rj~Skerrv-Ryan, et~al.,
\newblock ``Natural tts synthesis by conditioning wavenet on mel spectrogram
  predictions,''
\newblock in {\em 2018 IEEE International Conference on Acoustics, Speech and
  Signal Processing (ICASSP)}. IEEE, 2018, pp. 4779--4783.

\bibitem{lorenzo2018robust}
Jaime Lorenzo-Trueba, Thomas Drugman, Javier Latorre, Thomas Merritt, Bartosz
  Putrycz, Roberto Barra-Chicote, Alexis Moinet, and Vatsal Aggarwal,
\newblock ``Robust universal neural vocoding,''
\newblock {\em arXiv preprint arXiv:1811.06292}, 2018.

\bibitem{skerry2018towards}
RJ~Skerry-Ryan, Eric Battenberg, Ying Xiao, Yuxuan Wang, Daisy Stanton, Joel
  Shor, Ron~J Weiss, Rob Clark, and Rif~A Saurous,
\newblock ``Towards end-to-end prosody transfer for expressive speech synthesis
  with tacotron,''
\newblock {\em arXiv preprint arXiv:1803.09047}, 2018.

\bibitem{nachmani2018fitting}
Eliya Nachmani, Adam Polyak, Yaniv Taigman, and Lior Wolf,
\newblock ``Fitting new speakers based on a short untranscribed sample,''
\newblock {\em arXiv preprint arXiv:1802.06984}, 2018.

\bibitem{jia2018transfer}
Ye~Jia, Yu~Zhang, Ron Weiss, Quan Wang, Jonathan Shen, Fei Ren, Patrick Nguyen,
  Ruoming Pang, Ignacio~Lopez Moreno, Yonghui Wu, et~al.,
\newblock ``Transfer learning from speaker verification to multispeaker
  text-to-speech synthesis,''
\newblock in {\em Advances in neural information processing systems}, 2018, pp.
  4480--4490.

\bibitem{akuzawa2018expressive}
Kei Akuzawa, Yusuke Iwasawa, and Yutaka Matsuo,
\newblock ``Expressive speech synthesis via modeling expressions with
  variational autoencoder,''
\newblock {\em arXiv preprint arXiv:1804.02135}, 2018.

\bibitem{lorenzo2018investigating}
Jaime Lorenzo-Trueba, Gustav~Eje Henter, Shinji Takaki, Junichi Yamagishi,
  Yosuke Morino, and Yuta Ochiai,
\newblock ``Investigating different representations for modeling and
  controlling multiple emotions in dnn-based speech synthesis,''
\newblock {\em Speech Communication}, vol. 99, pp. 135--143, 2018.

\bibitem{wu2018rapid}
Xixin Wu, Yuewen Cao, Mu~Wang, Songxiang Liu, Shiyin Kang, Zhiyong Wu, Xunying
  Liu, Dan Su, Dong Yu, and Helen Meng,
\newblock ``Rapid style adaptation using residual error embedding for
  expressive speech synthesis.,''
\newblock in {\em Interspeech}, 2018, pp. 3072--3076.

\bibitem{gao2018nonparallel}
Jian Gao, Deep Chakraborty, Hamidou Tembine, and Olaitan Olaleye,
\newblock ``Nonparallel emotional speech conversion,''
\newblock {\em arXiv preprint arXiv:1811.01174}, 2018.

\bibitem{shankar2019multi}
Ravi Shankar, Jacob Sager, and Archana Venkataraman,
\newblock ``A multi-speaker emotion morphing model using highway networks and
  maximum likelihood objective,''
\newblock {\em Proc. Interspeech 2019}, pp. 2848--2852, 2019.

\bibitem{prateek2019other}
Nishant Prateek, Mateusz {\L}ajszczak, Roberto Barra-Chicote, Thomas Drugman,
  Jaime Lorenzo-Trueba, Thomas Merritt, Srikanth Ronanki, and Trevor Wood,
\newblock ``In other news: A bi-style text-to-speech model for synthesizing
  newscaster voice with limited data,''
\newblock {\em arXiv preprint arXiv:1904.02790}, 2019.

\bibitem{kingma2013auto}
Diederik~P Kingma and Max Welling,
\newblock ``Auto-encoding variational bayes,''
\newblock {\em arXiv preprint arXiv:1312.6114}, 2013.

\bibitem{tomczak2016improving}
Jakub~M Tomczak and Max Welling,
\newblock ``Improving variational auto-encoders using householder flow,''
\newblock {\em arXiv preprint arXiv:1611.09630}, 2016.

\bibitem{latorre2019effect}
Javier Latorre, Jakub Lachowicz, Jaime Lorenzo-Trueba, Thomas Merritt, Thomas
  Drugman, Srikanth Ronanki, and Viacheslav Klimkov,
\newblock ``Effect of data reduction on sequence-to-sequence neural tts,''
\newblock in {\em ICASSP 2019-2019 IEEE International Conference on Acoustics,
  Speech and Signal Processing (ICASSP)}. IEEE, 2019, pp. 7075--7079.

\bibitem{veaux2017cstr}
Christophe Veaux, Junichi Yamagishi, Kirsten MacDonald, et~al.,
\newblock ``Cstr vctk corpus: English multi-speaker corpus for cstr voice
  cloning toolkit,''
\newblock {\em University of Edinburgh. The Centre for Speech Technology
  Research (CSTR)}, 2017.

\bibitem{kominek2003cmu}
John Kominek, Alan~W Black, and Ver Ver,
\newblock ``Cmu arctic databases for speech synthesis,''
\newblock 2003.

\end{thebibliography}

\end{document}